\documentclass[twoside]{article} 
\pdfoutput=1
\usepackage{PRIMEarxiv}

\usepackage[utf8]{inputenc} 
\usepackage[T1]{fontenc}    
\usepackage{hyperref}       
\usepackage{url}            
\usepackage{booktabs}       
\usepackage{amsfonts}       
\usepackage{nicefrac}       
\usepackage{microtype}      
\usepackage{lipsum}
\usepackage{fancyhdr}       
\usepackage{graphicx}       

\usepackage{caption}
\usepackage{subcaption}
\usepackage{pifont}
\usepackage{multirow}
\usepackage{times}
\usepackage{epsfig}
\usepackage{amsmath}
\usepackage{amssymb}

\usepackage{amsmath,epsfig,bm, amssymb,graphicx,algorithm,algorithmic,color,url,multirow}

\newcommand{\mathbbR}{\mathbb{R}}

\newcommand{\boldX}{{\boldsymbol{X}}}

\newcommand{\bolds}{{\boldsymbol{s}}}

\newcommand{\ntr}{n}

\newcommand{\cmark}{\ding{51}}%
\newcommand{\xmark}{\ding{55}}%
\newcommand{\sizeformat}{0.5}%
\newcommand{\sizeformatsdcl}{0.5}%

\title{Scale dependant layer for \\ self-supervised nuclei encoding
\thanks{\textit{\underline{Citation}}: 
\textbf{Peter Naylor, Yao-Hung Hubert Tsai, Marick Laé and Makoto Yamada. Scale dependant layer for self-supervised nuclei encoding. ArXiv 2022. DOI:000000/11111.}} 
}

\newcommand{\authorspace}{\hspace{0.7cm}}%

\author{
    \authorspace \textbf{Peter Naylor} \authorspace\\
\authorspace RIKEN AIP \authorspace\\
\authorspace Kyoto, Japan \authorspace\\
\authorspace {\tt\small peter.naylor@riken.jp} \authorspace
\and
\authorspace \textbf{Yao-Hung Hubert Tsai} \authorspace\\
\authorspace Carnegie Mellon University \authorspace\\
\authorspace Pittsburgh, Pennsylvania, USA\authorspace\\
\authorspace {\tt\small yaohungt@cs.cmu.edu } \authorspace

\and
\\\authorspace \textbf{Marick Laé} \authorspace\\
\authorspace Centre Henri Becquerel - Unicancer \authorspace\\
\authorspace Rouen, France \authorspace\\
\authorspace {\tt\small  marick.lae@chb.unicancer.fr} \authorspace
\and

\\\authorspace \textbf{Makoto Yamada} \authorspace\\
\authorspace RIKEN AIP -- Kyoto University \authorspace\\
\authorspace Kyoto, Japan\authorspace \\
\authorspace {\tt\small makoto.yamada@riken.jp} \authorspace
}

\newcommand*{\inserttnbcconsepsamples}{
    \begin{figure}
        \centering
        \begin{subfigure}[b]{0.22\textwidth}
            \centering
            \includegraphics[width=1.\textwidth]{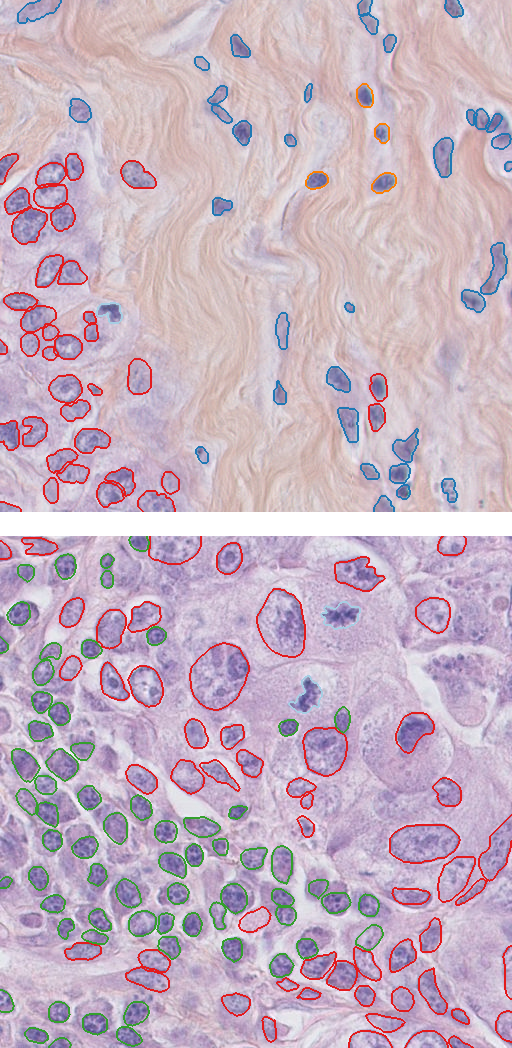}
            \caption{TNBC}
            \label{fig:tnbc}
        \end{subfigure}
        \begin{subfigure}[b]{0.22\textwidth}
            \centering
            \includegraphics[width=1.\textwidth]{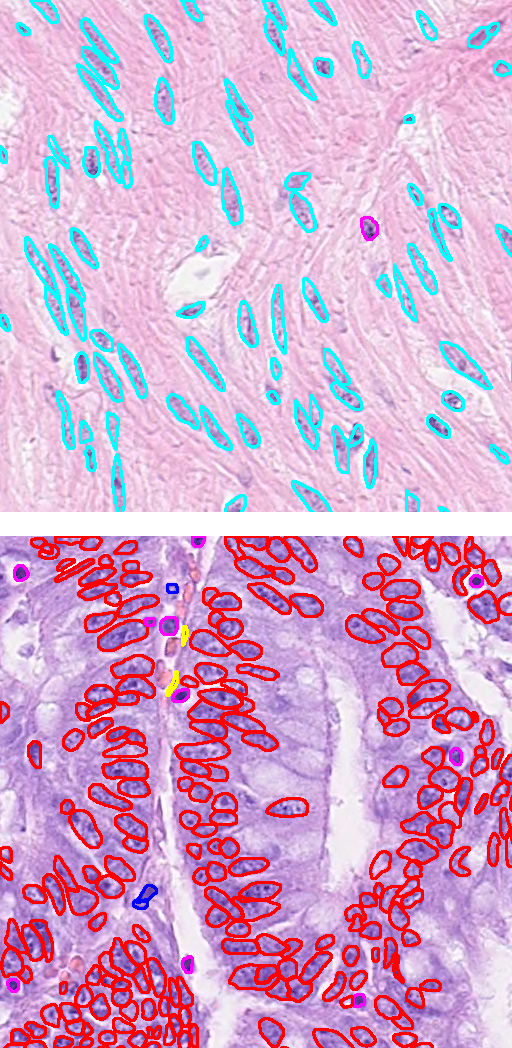}
            \caption{ConSep}
            \label{fig:consep}
        \end{subfigure}
        \caption{Dataset samples (in columns) with overlaid semantic segmentation.}
        \label{fig:dataset}
    \end{figure}
}

\newcommand*{\insertcelltypefigure}{
    \begin{figure}
        \centering
        \begin{subfigure}[b]{0.15\textwidth}
            \centering
            \includegraphics[width=0.8\textwidth]{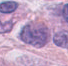}
            \caption{Cancerous}
            \label{fig:cancer}
        \end{subfigure}
        \begin{subfigure}[b]{0.15\textwidth}
            \centering
            \includegraphics[width=0.8\textwidth]{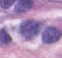}
            \caption{Lymphocyte}
            \label{fig:lympho}
        \end{subfigure}
        \begin{subfigure}[b]{0.15\textwidth}
            \centering
            \includegraphics[width=0.8\textwidth]{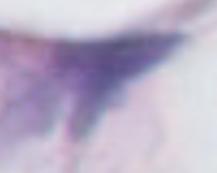}
            \caption{Adipocyte}
            \label{fig:adipocyte}
        \end{subfigure}
        \caption{Nucleus type example}
        \label{fig:three nuclei}
        \vspace{-.15in}
    \end{figure}
}

\newcommand*{\insertweighs}{
    \begin{figure}
        \centering
        \begin{subfigure}[b]{0.2\textwidth}
            \centering
            \includegraphics[width=0.9\textwidth]{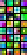}
            \caption{Standard convolution}
            \label{fig:sresnetw}
        \end{subfigure}
        \begin{subfigure}[b]{0.2\textwidth}
            \centering
            \includegraphics[width=0.9\textwidth]{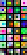}
            \caption{SD-CL}
            \label{fig:sdclw}
        \end{subfigure}
        \caption{Visualisation of the weights learned weights for the supervised model trained on PanNuke.}
        \label{fig:weights}
        \vspace{-.15in}
    \end{figure}
}

\begin{document}

\maketitle

\begin{abstract}
   Recent developments in self-supervised learning give us the possibility to further reduce human intervention in multi-step pipelines where the focus evolves around particular objects of interest.
   In the present paper, the focus lays in the nuclei in histopathology images.
   In particular we aim at extracting cellular information in an unsupervised manner for a downstream task.
   As nuclei present themselves in a variety of sizes, we propose a new Scale-dependant convolutional layer to bypass scaling issues when resizing nuclei.
   On three nuclei datasets, we benchmark the following methods: handcrafted, pre-trained ResNet, supervised ResNet and self-supervised features.
   We show that the proposed convolution layer boosts performance and that this layer combined with Barlows-Twins allows for better nuclei encoding compared to the supervised paradigm in the low sample setting and outperforms all other proposed unsupervised methods.
   In addition, we extend the existing TNBC dataset to incorporate nuclei class annotation in order to enrich and publicly release a small sample setting dataset for nuclei segmentation and classification.
\end{abstract}

\section{Introduction}
Histopathology is the study of diseased tissue under a microscope, these image data correspond to tissue slides encompassing the tumour and the surrounding tissue.
Digital pathology emerged in 60's and since, supported clinicians and informatics pipelines for the analysis of histopathology samples.
Prior to 2012, automatic algorithms and classical machine learning (in contrast to deep learning) have left the field relatively untapped.
Due to the inherent complexity of histopathology specimens, most studies were usually highly specific and performed on homogeneous small cohorts \cite{irshad2013methods}.
The advent of deep learning, where automated algorithms reached unprecedented results in computer vision, have fuelled investments and research in this field \cite{kaushal2019recent}.
The field of digital pathology topical has become topical through the ever growing number of challenges, scientific papers and publicly available datasets \cite{litjens20181399, aresta2019bach, caicedo2019nucleus, graham2021lizard}.

\begin{figure}[t]
  \centering
  \includegraphics[keepaspectratio,width=1.0\linewidth]{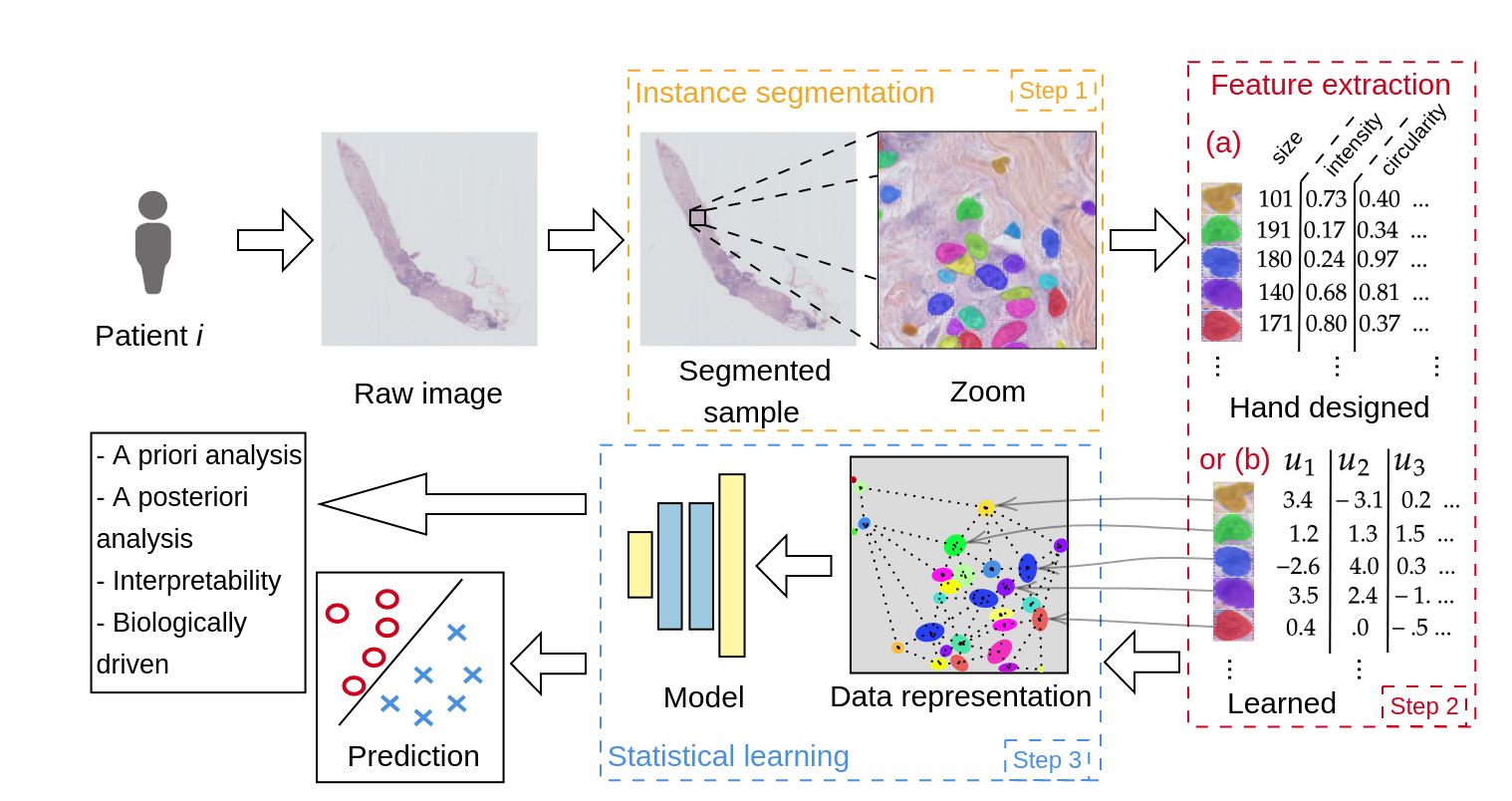}
  \caption{Feature extraction pipeline. We first performed instance segmentation, followed by feature extraction for each object. In the present article, we focused on step 2.\label{fig:pipeline}}
\end{figure}

Histopathology studies can be divided into ``basic'' and ``advanced'' tasks \cite{echle2021deep}.
Basic tasks range from tumour detection and grading to sub-type classification, i.e. relatively known and formalized tasks.
For these tasks deep learning research is mostly oriented to help pathologists in their every day work by simplifying workflows \cite{shvetsov2022pragmatic}.
In contrast, advanced tasks range from survival, mutation to treatment response prediction and go beyond the usual reporting asked from pathologists.
In order to leverage the power of deep learning in the field of histopathology, it is essential to build explainable and interpretable models that study biological elements of interest.
At the moment, there are two classes of algorithms: those that aim at predicting clinical variables directly from the image data \cite{dehaene2020self}, and those which seek to profile the images in terms of biologically meaningful features \cite{neumann2010phenotypic}.
These latter algorithms involve, typically, the segmentation of an object of interest, like the nuclei.
Then a quantitative description of their phenotypes and context, i.e. density information, morphological properties and spatial disposition. 
This step is followed by specimen classification based on the previous extracted attributes.
The introduction of this intermediate step is the most logical way of imposing biological interpretability as the tissue slides are described in terms of cellular and tissular phenotypes \cite{neumann2010phenotypic, mcquin2018cellprofiler}. 
In Fig. \ref{fig:pipeline} we show an example of such a pipeline.

This study fits in a larger framework presented in Fig. \ref{fig:pipeline} and we focus on the second step in the present article.
In particular, we study the encoding of a meaningful element in histopathology analysis, the nuclei.
The nuclei is the most important element, and is the most commonly studied element in the field  \cite{neumann2010phenotypic, held2010cellcognition, irshad2013methods, mcquin2018cellprofiler, caicedo2019nucleus}.
The nuclei class and status, as well as its abundance, hold valuable information for the elaboration of diagnosis and treatment for many diseases \cite{symmans2007measurement}.
Therefore building a multi-step model where we boil down the analysis through the nuclei is highly relevant.
If successful, these multi-step models allow for apriori, via the segmentation and quantification of nuclei before sample prediction, and aposteriori analysis, by back-propagating the prediction back to the nuclei.
Such analysis can be a powerful tool for assisting current and fuelling future biological and medical research. 

For the second step, there exist accurate nuclei detection algorithms that have been extended to Whole Slide Imaging \cite{graham2019hover, amgad2021nucls}, and thus the success of the pipeline is to encode nuclei accurately.
In the past decades, researchers have developed useful manual features for classifying nuclei and use machine learning algorithms such as the support vector machines (SVM) \cite{drucker1996support,neumann2010phenotypic}.
However, designing manual features is time-consuming, error prone and sub-optimal for a given task. 
An alternative approach is to use deep learning models such as the residual networks \cite{he2016deep} where end-to-end learning is adopted and the feature extractor module is optimized according to a given task.  
However, histopathology data, depending on the tissue and disease, presents a fairly heterogeneous collection of cells, i.e., the size, morphology and texture of each detected nuclei is different. 
More specifically, cell type also depends on the context and the organ in which the cell is present.
When describing nuclei and visualizing them, we do not expect clearly distinguishable clusters but rather a continuum between cell class due to the wide range and variability found in their phenotypes. 

Moreover, since only trained researchers/technicians can annotate nuclei classes, it is challenging to create a large scaled dataset for nuclei classification. 
One approach is to use machine learning algorithms to annotate nuclei images \cite{gamper2019pannuke,amgad2021nucls,graham2021lizard}, however, annotations are usually restricted to a small number of classes and do not cover the whole spectrum of nuclei classes.
Indeed for every tissue type, nuclei class can be refined into many more classes \cite{gamper2019pannuke}. 
Thus, there exist a small number of publicly available accurate annotated datasets for nuclei classification.


In this paper, to address the heterogeneity in nuclei size, we propose the Scale Dependant dilation convolution.
The proposed operation takes in input a resized object and the original object size to allow for a meaningful and comparable feature computation from different image scales.
Thus this layer allows different scales to be combined while retaining the advantages of a homogeneous image size, highly valued in deep neural networks.
Then to study the small sample case, we extended the Triple Negative Breast Cancer (TNBC) dataset \cite{naylor2018segmentation} by adding cell type to the dataset.
This is, currently, the only hand segmented and annotated TNBC available dataset.
Moreover, we propose a self-supervised training framework for nuclei classification with a benchmark of methods.
Through experiments, we demonstrate that our proposed Scale Dependant convolution layer achieves better results with the Barlows-Twins (BT) compared to manual models, supervised and other state of the art self-supervised models.

The contribution of this paper is summarized as follows:
\begin{itemize}
   \item We extend the TNBC dataset by adding cell type to the dataset. In particular, $4,056$ nuclei are annotated in $9$ classes.
   \item A scale dependant convolution layer that can be used to extract features from objects at different scales.
   \item We demonstrate the utility of the scale dependant layer by conducting a benchmark on three datasets that shows that it has a positive impact the performance.
   In particular, we compare manual, pre-trained, self-supervised and supervised features.
\end{itemize}



\section{Related work}
In this section, we review existing datasets for nuclei classification, feature extraction, and scale invariant methods, respectively. 
Then, we discuss the potential problems in the current nuclei classification methods.

\vspace{.05in}
\noindent {\bf Annotated nuclei datasets:}
Due to its essential role in the automatic interpretation of stained tissue sections, segmentation of nuclei has been addressed by many authors with a variety of traditional approaches, see also \cite{irshad2013methods} for an extensive reviews. 
The aim of this field is to quantify by nuclei class the tumour micro environment, and to use this information as a profiling tool, diagnosis or in-depth analysis of tumour subtypes \cite{ahmedt2021survey,jaume2021quantifying}
Today, deep neural networks are considered the best tool for this task \cite{naylor2018segmentation,lagree2021review,graham2021lizard}. 
These models, based on supervised learning, come hand in hand with manually annotated datasets, and depending on the number of annotators, typically range in the thousands to tens of thousands of annotated nuclei \cite{graham2021lizard}. 
Lately, datasets have reached hundreds of thousands of annotated nuclei \cite{amgad2021nucls, graham2021lizard} with the help of automated algorithms.
They propose a data generation pipeline where they leverage an already trained network to generate segmentation for unseen data.
The new data is then checked and iteratively refined through semi-automatic and manual procedures.
We summarize other existing datasets in Table \ref{nuc_data}, in this work we extend the TNBC dataset by including cell annotation.

\begin{table}
    \begin{center}
        \resizebox{\sizeformat\linewidth}{!}{
            \begin{tabular}{c|c|c|c|c}
        
            \hline   
            Dataset & Nuclei  & Labels & Organs   & Semi- \\
                    &         &        &          & automatic \\
            \hline 
            \hline 
            TNBC  \cite{naylor2018segmentation}  & 4,056   & \cmark & Breast   &  \xmark  \\
            \hline 
            CPM-17 \cite{Kurc2018Miccai}  & 7,570   & \xmark & Multiple &  \xmark          \\
            \hline 
            MoNuSeg \cite{kumar2019multi} & 21,623  & \xmark & Multiple &    \xmark  \\
            \hline 
            CoNSep \cite{graham2019hover} & 24,319  & \cmark & Colon    &   \xmark  \\
            \hline 
            MoNuSac \cite{verma2020multi} & 46,909  & \cmark & Multiple &  \xmark  \\
            \hline 
            PanNuke \cite{gamper2019pannuke} & 189,744 & \cmark & Multiple &  \cmark   \\
            \hline 
            NuCLS \cite{amgad2021nucls}  & 222,396 & \cmark & Breast   &  \cmark    \\
            \hline 
            Lizard \cite{graham2021lizard} & 495,179 & \cmark & Colon    &  \cmark    \\
            \hline 
            \end{tabular}
            }
            \caption{Nuclei segmentation datasets.\label{nuc_data}}
    \end{center}
    \vspace{-.15in}
 \end{table}

The current trend is to build models like the HoVer-Net \cite{graham2019hover} where segmentation and cell class prediction is done in a single pass.
The work presented in the current article differs from these in two manners. 
Firstly, we aim to build an unsupervised feature extraction pipeline to allow the quantification of tissue that does not rely on hard nuclei labelling but rather on a soft representation.
Secondly, because the spectrum of nuclei class is huge and dependant on the tissue and disease, we aim to quantify every nuclei in a general manner to not restrict the downstream analysis to fixed labels but rather to these soft representations that allow the whole spectrum of nuclei classes to be represented.

Moreover, cell types prediction can be challenging.
In Hematoxylin and Eosin (H\&E) staining, similar nuclei morphologies does not necessarily imply same class.
For instance, epithelial cells can appear small and dense like lymphocytes.
Glial cells found in brain tissue are also similar to lymphocytes.
Macrophages can be mistaken for other cells when they are digesting and cell type is hard to distinguish for mitotic events, in these situations context can shed information about the type.


\vspace{.05in}
\noindent {\bf Feature extraction:}
Image encoding or representation learning in its most primal state can be a shallow classifier where the prediction is based on a continuous hidden variable, such as the distance to the hyperplane.
The underlying hope is that this hidden variable acts as a compact representation from the original high-dimensional data.
Prior to deep learning, hand designed features were commonly used in computer vision. 
In particular, in image analysis pipelines where one had to manually design a feature extractor for an underlying task, including SIFT~\cite{lowe2004distinctive} and SURF~\cite{bay2006surf}. 
Now, the scientific community has moved to deep networks to learn non-manual-designed features. 
The first famous example was trained on ImageNet \cite{deng2009imagenet} and this dataset is still used today for pre-training the networks.
Indeed the large number of classes forces the feature extractor to cope with a large heterogeneity of images and labels making it general.
Pre-training has mostly been used when the number of samples is low, even when ImageNet and the target dataset differ in a number of ways \cite{raghu2019transfusion}.

Training from scratch, i.e. end to end training, is a means to overcome the biases of the pre-trained network.
In particular, recent advances show that self-supervised learned features perform on a par with supervised learned features \cite{chen2020simple,misra2020self}. 
We can categorize self-supervised learning approaches into three groups: contrastive learning approaches~\cite{he2020momentum,chen2020simple}, predictive learning approaches~\cite{he2022masked}, and others~\cite{zbontar2021barlow,grill2020bootstrap}.
For the first two groups: contrastive learning approaches aim to learn similar features for positively-paired samples and dissimilar features for negatively-paired samples, and predictive learning approaches aim to reconstruct a partial of inputs given the rest part of the inputs.
In this paper, we consider Barlows-Twins \cite{zbontar2021barlow} (BT) method, which considers to maximize cross-correlation matrix between augmented batch of inputs, as our self-supervised learning approach, and it belongs to the ``others'' category in self-supervised learning.

For histopathology data, self-supervised learning has been used in multi-step approaches to classify large images as end-to-end learning is impractical.
The multi-step approach consists of encoding tiles followed by the classification of the whole image \cite{dehaene2020self,lazard2021deep,naylorprediction}.
Prior to self-supervised learning, the most common approach was to use pre-trained networks on ImageNet to encode the tiles \cite{raghu2019transfusion}.
To the best of our knowledge, only one other paper has studied the encoding of nuclei in histopathology data \cite{graham2021lizard}, which uses contrastive learning for nuclei representation.
To account for the difference in size they crop with a large window, and mask the surrounding during training.
They report the Cohen's kappa agreement measure between trained pathologist and their prediction, found via hierarchical clustering of the contrastive representation.
However no accuracy measure on the classification of nuclei is reported.


\vspace{.05in}
\noindent {\bf Scale invariant neural network:}
The most straightforward manner to impose invariance in a neural network is to randomly augment the input data.
For scale invariance, input images are randomly rescaled in order to render the network invariant to the scale of the object to a certain degree. 
To address the topic further, two approaches have been adopted.
The first involves building models inherently invariant to scaling \cite{xu2014scale, kanazawa2014locally} or by using invariance properties to constrain the parameters.
For instance, multi-branch networks have been proposed, where each branch has its own filter and scale, leading to a model where different scales are handle and combined for the final prediction \cite{xu2014scale}.
Multi-scale pyramidal models with skip layers to allow the combination of large and small scale information have been successful \cite{ronneberger2015u, liu2019scale} and used for general object detection tasks with the RCNN \cite{ren2015faster}.
Moreover, invariances can induce symmetries that are efficient methods to reduce the number of parameters \cite{kanazawa2014locally}.
The second type of approach focuses on leveraging a previously successful scale invariant model to become scale aware \cite{graziani2020interpretable,graziani2021scale}.

Dilated convolutions, and more generally deformable convolutions  \cite{yu2015multi, dai2017deformable, mumuni2021cnn, hassani2021dilated} allow a model to better adapt to the local geometry found in the dataset.
In particular, it can adapt itself by finding the best window, or scale for image feature extraction.
The work we present in this article uses parametrized deformable convolution, and in contrast to previous work, does not aim to learn the best spacing.
Specifically, we see the dilated convolution as a parametrize and input-dependant function.
The aim is to extract features while preserving the initial shape of the object by using deformable convolutions.

\section{Dataset} \label{dataset}
We introduce the datasets used in this paper. 
Moreover, we extend the TNBC dataset by adding cell type annotation to the existing dataset.

\vspace{.05in}
\noindent {\bf TNBC:} The Triple Negative Breast Cancer (TNBC) dataset published by the Curie Institute consists of 50 annotated H\&E stained histology images at $40\times$ magnification.
In particular $4,056$ nuclei boundaries were annotated.
This dataset has been complemented and the nuclei types are now also publicly available, see Fig. \ref{fig:tnbc} for some samples.
Each nuclei was annotated into one of the $9$ classes. 
The nuclei state mitosis is not a true nuclei type per say but a nuclei state.
We decided on adding this notation as the localisation and detection of mitosis is an important topic and is involved in quantifying cancer growth. 
In addition, any nuclei type can undergo mitosis and can look morphologically similar.
The annotation is summarized in Table \ref{tab:TNBC}.

\begin{table}
    \begin{center}
        \resizebox{1.0\linewidth}{!}{
            \begin{tabular}{l|rrrrrrrrrrr|r}
            \hline
            Slide number &  01 &  02 &  03 &  04 &  05 &  06 &  07 &  08 &  09 &  10 &  11 &  Total \\
            \hline
            Number of patches & 7 & 3 & 5 & 8 & 4 & 3 & 3 & 4 & 6 & 4 & 3 &  50 \\
            \hline
            \hline
            Adipocyte &        0 &        1 &        0 &       32 &        0 &        0 &        0 &        0 &        5 &        1 &        0 &       39 \\
            Cancerous &      243 &      150 &      133 &      443 &      306 &      110 &       48 &      187 &      178 &      192 &      141 &    2131 \\
            Endothelial &      12 &        0 &        0 &        7 &        0 &        3 &        6 &       35 &        3 &        7 &        8 &      81 \\
            Epithelial &      22 &        0 &        0 &        0 &        0 &        0 &       34 &        0 &        0 &        0 &        0 &        56 \\
            Fibroblast &      136 &        8 &       95 &       92 &      105 &       85 &      155 &        5 &        1 &        9 &       97 &     788 \\
        Lympho/Plasmocyte&       43 &       29 &       58 &       20 &        8 &        6 &      307 &      212 &      103 &      105 &        6 &      897 \\
            Mitosis    &        2 &        0 &        1 &        2 &        0 &        0 &        0 &        6 &       11 &        2 &        2 &        26 \\
            Myoepithelial  &  0 &        0 &        0 &        0 &        0 &        0 &        4 &        0 &        0 &        0 &        0 &          4 \\
            Necrosis   &        0 &        0 &        0 &        0 &        0 &        0 &        1 &        0 &        0 &        0 &        0 &        1 \\
            \hline
            Total     &      458 &      188 &      287 &      596 &      419 &      204 &      555 &      445 &      301 &      316 &      254 &     5150 \\
            \hline
            \end{tabular}%
 }\caption{Detailed account of nuclei type within each sample of the TNBC \cite{naylor2018segmentation} dataset.\label{tab:TNBC}}
    \end{center}
    \vspace{-.15in}
 \end{table}

The nuclei type annotations were added via the software CellCognition \cite{held2010cellcognition}.
CellCognition is usually used for annotating fluorescent microscopy images and can also be used for any image and object type annotation.
With this tool, we annotated cell types which were all checked by an expert pathologist.


\vspace{.05in}
\noindent {\bf  CoNSep:} The colorectal nuclear segmentation and phenotypes dataset (CoNSep) published by the department of pathology at University Hospitals Coventry and Warwickshire consists of 41 annotated H\&E stained histology images at $40\times$ magnification.
In particular $24,909$ nuclei boundaries and nuclei types were annotated into 7 classes, see Fig. \ref{fig:consep} for some samples.
As with the original paper, we combine classes 3 and 4  into the epithelial class and 5,6 and 7 into the spindle-shaped class.

\inserttnbcconsepsamples

\vspace{.05in}
\noindent {\bf PanNuke:} The PanNuke \cite{gamper2019pannuke} dataset was also published by the department of pathology at University Hospitals Coventry and Warwickshire.
It consists of $189,744$ annotated nuclei in a semi-automated manner from multiple tissue divided into 5 classes.


\vspace{.05in}
\noindent {\bf Dataset preparation and manual feature extraction:} For each nucleus, we extract a total of $68$ features. 
We measure the nucleus size, width, height, elongation, circularity.
We add color information by measuring the average and standard deviation of the intensity of the nuclei on each color bandwidth and its greyscale transformation.
To include edge information as well as texture information, we include local binary patterns (LBP) \cite{ojala2002multiresolution} and granulometric features \cite{chen1994gray}.
LBP are operators that extract edge information by sliding a window along the image, this results into a distribution that we numerically quantify with quantiles ranging from $10\%$ to $90\%$, moreover we apply this operator with a radius equal to $1$ and $3$ pixels.
We apply granulometric features on the greyscale image at sizes ranging from $1$ to $5$. 
Finally, in order to allow the incorporation of surrounding information, we apply the same set of features to the dilated version of the nuclei. 
We use a dilation of factor of $4$ and we do not extract the width, height, elongation and circularity of the dilated nucleus.

Size information, shape, intensity as well as texture information is valuable when differentiating between cell types.
Lymphocyte tend to be smaller, darker and have a color distribution that is relatively homogeneous, see Fig. \ref{fig:cancer}.
Cancer nuclei are usually larger, irregularly shaped, with irregular intensities and tend to have a lighter color, see Fig. \ref{fig:lympho}.
Adipocyte, similarly to lymphocyte, tend to be small, darker and homogeneous.
They differ only in their shape and context, indeed, they are usually compressed by fat and are much more likely to be elongated, see Fig. \ref{fig:adipocyte}.

From this semantic segmentation dataset, we crop each connected component with a $5$ pixel margin on each side and resize all nuclei to a fixed $32\times$ $32$ size.

\insertcelltypefigure

\section{Proposed Method} \label{models}
We will compare the previously described manual features to learned representation.
In particular, to a pre-trained ResNet on ImageNet, to a benchmark of supervised models and self-supervised models, MoCo \cite{he2020momentum} and BT \cite{zbontar2021barlow}.

\subsection{Problem formulation}
Let us denote an input image by $\boldX_i \in \mathbbR^{h_i \times w_i \times 3}$ and  the corresponding output class labels $y \in \{1,2,\ldots, L\}$, where the image input size is sample dependant, i.e the height $h_i$ and the width $w_i$ can be different and $L$ is the total number of classes.
For nuclei data, as the size of the nuclei are different, this setup is reasonable. 
Then, we resize $\boldX_i$ into $\widetilde{\boldX_i}\in \mathbbR^{32 \times 32 \times 3}$ and construct the set of samples $\{(\widetilde{\boldX_i}, \bolds_i, y_i)\}_{i = 1}^{\ntr}$, where $\bolds_i = (h_i, w_i)^\top \in \mathbbR_+^2$ is the size information of the original image $\boldX_i$.

The goal of this paper is to build meaningful nuclei encoding that represent the wide heterogeneity found in histopathology data.
We check the efficiency of the encoding via two steps
In the first, we encode the samples via one of the proposed methods. 
Then, we apply a classification model on top of the encoding and report two metrics.
Specifically, we use relatively simple classification models such as a single layer neural network and a nearest neighbour algorithm.

\subsection{Backbone model}
The supervised and self-supervised model both rely on a backbone model.
We implement a modified and adapted version of the 34-layer ResNet for smaller images, named SResNet.
In particular we replace the initial $7\times7$ and 64 feature convolutional block and max-pooling layer by a $3\times3$ and 32 features convolutional layer with batch normalisation and ReLu activation.
We then stack three residual blocks consisting of $3$, $4$ and $3$ residual layers each and $32$, $64$ and $128$ features respectively.
We reduce the size of the feature maps with a striding of two after the second and third residual blocks.
We then use an average pooling layer to reduce the spatial resolution to $1$ to produce the image encoding.
In the supervised setting, we apply a final fully connected layer to produce a probability vector with respect to the number of classes.

\subsection{Dealing with resizing}
\begin{figure}[t]
  \centering
  \includegraphics[keepaspectratio,width=1.0\linewidth]{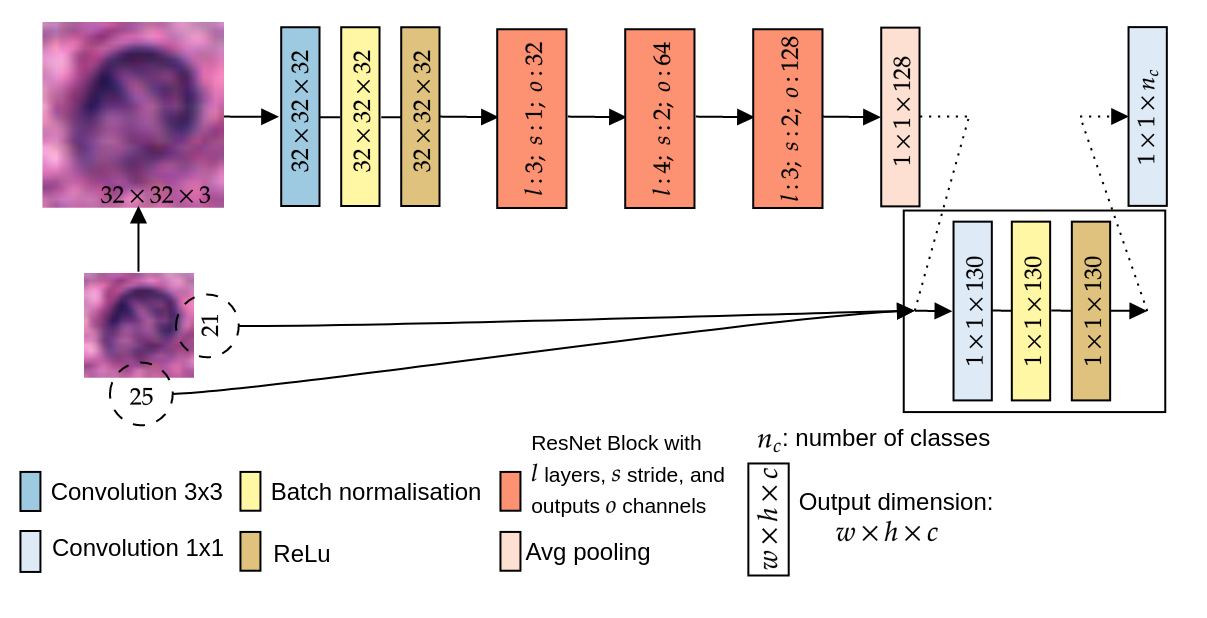}
  \caption{SResNet with size concatenation.\label{fig:backbone}}
\end{figure}

\vspace{.05in} \noindent {\bf Size injection:}
As we presume that size information is valuable and absent from the backbone, we simply inject the original width and height of the input image to the encoding in the latter layers of the network.
Moreover, in order to allow the backbone to adjust the representation, we add an extra fully connected layer with batch normalization and ReLu.
In particular the layer prior to the  final probability layer is the new encoding, see Fig. \ref{fig:backbone}.

\vspace{.05in} \noindent {\bf Adjustable dilation:}
The main issue with resizing the images prior to their analysis happens in the first layer.
Indeed, if the images are resized, we are extracting and comparing features from different scales.
However, resizing allows for computational efficiently and batching of the samples.
Therefore all input are resized to a fixed $32\times32$ size.

In the current backbone, the first convolutional layer extract and compares features that ultimately have different scales.
This issue is more important for the first layers where the networks learn basic features, such as edges, corners and color gradients.
We oppose basic features to more abstract features as they have been defined in other works \cite{zeiler2014visualizing}, it has been empirically shown that latter layers hold more abstract concepts.
We decide to use the original height and width to adjust the dilation in the first layer to compensate for the different scales and allow for a fairer and more scale invariant representation in the first layer.
We do not need to adjust the scale in the following layers as the inner representation should already be invariant to the scale.

We give below the equation for our parametrized dilated convolution for computing the resulting feature map $m$ at pixel $(i_0, j_0)$ for a one channel input image:
$$f_m^{sd}(I)[i_0, j_0] = \hspace{-0.4cm} \sum_{i, j=-k_w, -k_h}^{k_w, k_h} \hspace{-0.4cm}  I[i_0 + i * d_w(I), j_0 + j * d_h(I)] \times W_{i, j},$$
where $W$ is the learnable kernel with width $w_k$ and height $h_k$.
For the sake of clarity and conciseness, we assume the height and width of $W$ are odd and measured from the center pixel $W_{0,0}$.
For a $3\times3$ kernel, we have $w_k = h_k = 1$.
In particular we define $d_a(I) = \text{max}(\lfloor u_f \times 32 / a_I \rfloor, 1)$ which computes the dilation factor over axis $a$ (equal to the height or width axis), $a_I$ corresponds to the original length of $I$ over axis $a$.
$u_f$ corresponds to the up-scaling factor applied to the image $I$ prior to the convolution, set to $3$. 
In larger images, the up-scaling factors allow to reduce the dilation factor further.
Finally, max-pooling is applied to reduce the image back to the original size of the feature map, we give below the whole function:
$$F^{sd}(I) = \text{max-pooling}_3 \circ f^{sd} \circ \text{Upsample}_3 (I),$$
where $\text{Upsample}_3$ up-scales by factor $3$ and $\text{max-pooling}_3$ downsizes by a factor $3$ with a max-pooling operation.

We name this layer, SD-CL, for Scale Dependant Convolutional Layer.
With no loss of generality, we can also apply SD-CL to 3D objects.

\begin{figure}[t]
  \centering
  \includegraphics[keepaspectratio,width=\sizeformatsdcl\linewidth]{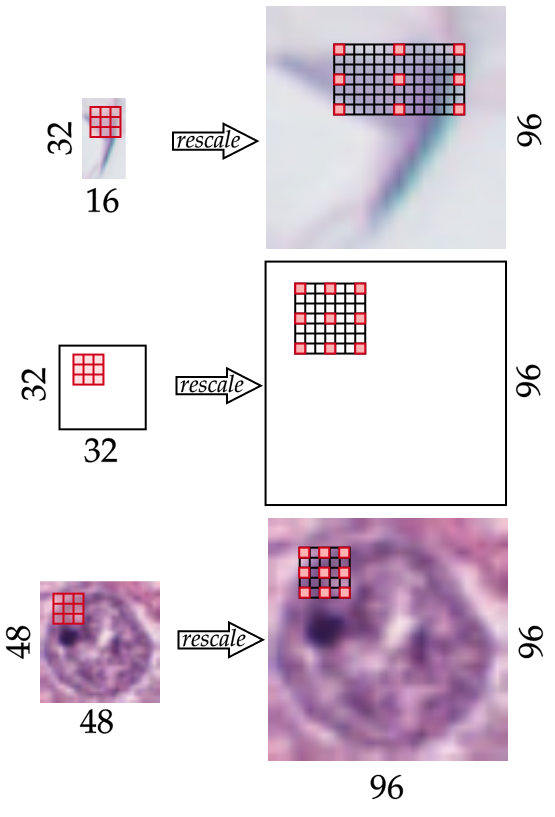}
  \caption{SD-CL: Scale Dependant convolutional layer with adjustable dilation. The kernel weights are shared.\label{fig:siconv}}
\end{figure}


\subsection{Models}

\vspace{.05in} \noindent {\bf Pre-trained model: } We encode each nuclei with a pre-trained ResNet on ImageNet.
In particular, for each image, we resize it to $224 \times 224$ and extract the $2048$ encoding.

\vspace{.05in} \noindent {\bf Baseline supervised method:} As another baseline approach for the learning paradigm we propose a supervised framework. 
We compare SResNet, with and without size concatenation to the same models with the SD-CL.


\vspace{.05in} \noindent {\bf Momentum Contrast (MoCo) \cite{he2020momentum}:} 
During training we infer queries and match these to a set of keys, that correspond to encoded samples.
In particular, it is assumed that there is a single key $k_+$ that matches a query $q$. 
We minimize the following loss:
$$\mathcal{L}_{q}=-\log \frac{\exp \left(q \cdot k_{+} / \tau\right)}{\sum_{i=0}^{K} \exp \left(q \cdot k_{i} / \tau\right)},$$
where $K$ denotes the number of negative samples and $\tau$ the temperature. 
During training, a positive match corresponds to a given image and its augmented version.
Negative keys correspond to the other encoded samples.

\vspace{.05in} \noindent {\bf Barlows-twins \cite{zbontar2021barlow}: } Barlows-twins is a self-supervised method, where given a model $f_{\theta}$ and set of random transformations (or data augmentation) function $\mathcal{T}$ we minimize the following loss $\mathcal{L}_{\mathcal{BT}}$:
$$\mathcal{L}_{\mathcal{BT}} = \sum_{i}\left(1-\mathcal{C}_{i i}\right)^{2}+\lambda \quad \sum_{i} \sum_{j \neq i} \mathcal{C}_{i j}{ }^{2},$$ 
where $\mathcal{C}_{i j} = \frac{\sum_{b} z_{b, i}^{A} z_{b, j}^{B}}{\sqrt{\sum_{b}\left(z_{b, i}^{A}\right)^{2}} \sqrt{\sum_{b}\left(z_{b, j}^{B}\right)^{2}}}$ and $z^A$ and $z^B$ are random transformations of the same input feed to the model $f_{\theta}$. 
$\lambda$ is a hyper-parameter to balance between invariance and the redundancy reduction term.
The underlying idea with contrastive learning is that two transformed input images should have a similar embedding in the feature space.
It was shown in the original paper that it is better when applying neural networks to add a projection head to the backbone.
Like the supervised paradigm, SResNet is used as the backbone and  we experiment with the SD-CL and size injection.

\section{Experiments and results} \label{experiments}

In this section, we evaluate our proposed algorithms using our newly constructed TNBC datasets and two other datasets, CoNSep and PanNuke.

\subsection{Experimental details}
\noindent {\bf Data splits:} For all datasets, we split the dataset into a training and test sets.
For TNBC, we use the samples folds 1 and 9, for the test set in order to have sufficient labelled samples in each class.
Moreover we remove the minor classes Myoepithelial and Necrosis, resulting in $3,259$ train samples and $759$ test samples.
For CoNSep, we use the split given by the authors that allows us to constitute a dataset of $15,555$ train samples and $8,777$ test samples.
Similarly for PanNuke, we use the pre-defined folds and use fold 1 and 2 for training and fold 3 for testing. 
For this dataset we have $122,117$ train samples  $67,627$ test samples.
During model selection and depending if the method requires tuning, we further split the training dataset into train and validation set in order to perform model selection in an unbiased manner.



\vspace{.05in}\noindent {\bf Training details:} To refine the manually designed features we perform feature selection in a similar fashion to the forward and backwards stepwise selection \cite{hocking1976biometrics}.
In the forward method, we add features if this improves the negative log-likelihood on the validation set.
Similarly, in the backward method, we remove features if this improves the negative log-likelihood minus some heuristic scalar (set to $0.05$) on the validation set.
Finally, we compare the forward selection, the backward selection, the intersection and the union of the selected features on the validation set and retain the best performing set of features.

For the learning paradigms we use an Adam optimizer and optimize the learning rate, weight decay and $\lambda$ for BT on the validation set.
We use a batch size of $128$ for TNBC and CoNSeP.
For PanNuke we use a batch size of $512$.
We set the number of epochs to $100$ except for PanNuke where the number of training iterations is reduced to $50$.
We use the following set of transformations, all applied with probability $p=0.5$: rotation, vertical and horizontal flip, color jitter, resized crop and greyscale transformations.
For the rotation and resized crop we modify the injected height and width accordingly.
In particular, a rotation of an angle $\theta$ leads to a new set of heights and widths defined as: 
\begin{align*}
   h_{\theta} & = h.\text{cos}\theta + w. \text{sin} \theta, ~~w_{\theta}  = h.\text{sin}\theta + w. \text{cos} \theta.
\end{align*}

\vspace{.05in}\noindent {\bf Metrics:} We use an accuracy score based on the training of a one layer Softmax classifier on top of the embedding given by the selected model, we name this metric the linear accuracy metric.
In addition, we use a nearest neighbour classifier with $k=50$ on the given encoding, we name this the kNN accuracy metric. 
We prefer to maximise the nearest neighbour score as this one is more relevant as it shows that the embedding space holds semantic information.
In particular, we select the model that maximises the nearest neighbour validation accuracy score.

\subsection{Application to nuclei encoding}

\begin{table*}[!ht]
   \begin{center}
      \resizebox{1.0\linewidth}{!}{
         \begin{tabular}{c|c|cc|cc|cc}
            \toprule
            \multicolumn{1}{c}{}&                              & \multicolumn{2}{c|}{TNBC}    & \multicolumn{2}{c|}{CoNSep}   & \multicolumn{2}{c}{PanNuke}   \\
            \cmidrule{2-8}
            \multicolumn{1}{c}{}&$L$ & \multicolumn{2}{c|}{7}   & \multicolumn{2}{c|}{4}        & \multicolumn{2}{c}{5 }   \\
            \multicolumn{1}{c}{}&Training set size             & \multicolumn{2}{c|}{$3,259$} & \multicolumn{2}{c|}{$15,555$} & \multicolumn{2}{c}{$122,117$}   \\
            \multicolumn{1}{c}{}&Test set size                 & \multicolumn{2}{c|}{$759$}   & \multicolumn{2}{c|}{$8,777$}  & \multicolumn{2}{c}{$67,627$}   \\
            \cmidrule{2-8}
            \multicolumn{1}{c}{}&Accuracy metric               & Linear  & kNN         & Linear  & kNN       & Linear  & kNN      \\
            \cmidrule{1-8}
            \multirow{3}{*}{\rotatebox{90}{\scriptsize{No training}}}&Manual                        & $55.5 \pm 3.7$  & $55.6 \pm 0.0$ & $52.8 \pm 10.7$ & $17.6 \pm 15.2$   & $43.5 \pm 5.8$  & $42.4 \pm 0.0$   \\
            &Pre-trained ResNet                                         & $55.6 \pm 0.0$ & $51.3 \pm 0.0$ & $37.1 \pm 1.1$  & $37.8 \pm 0.0$    & $42.3 \pm 0.1$  & $42.4 \pm 0.0$   \\
            &Pre-trained ResNet + Size     & $55.6 \pm 0.1$ & $20.0 \pm 0.0$ & $37.1 \pm 1.6$  & $37.9 \pm 0.0$    & $42.3 \pm 0.1$  & $17.1 \pm 0.0$   \\
            \cmidrule{1-8}
            \multirow{4}{*}{\rotatebox{90}{Supervised}}&   SRN  & $57.3 \pm 1.2$                   & $52.6 \pm 8.2$          & $73.9 \pm 0.7$           & $71.3 \pm 1.7$           & $71.1 \pm 0.4$           & $64.1 \pm 1.6$   \\
            &SRN + Size                                         & $58.1 \pm 1.4$                   & $39.0 \pm 3.5$          & $74.1 \pm 0.8$           & $58.6 \pm 2.5$           & $\mathbf{73.1 \pm 0.3}$  & $45.3 \pm 2.2$   \\
            &SRN + SD-CL                                        & $\mathbf{58.8 \pm 0.9}$          & $\mathbf{58.9 \pm 2.3}$ & $\mathbf{74.4 \pm 0.7}$  & $\mathbf{71.6 \pm 1.5}$  & $72.2 \pm 0.3$           & $\mathbf{64.9 \pm 1.6}$   \\
            &SRN + SD-CL + Size                                 & $58.4 \pm 0.9$                   & $46.7 \pm 3.4$          & $74.0 \pm 0.4$           & $61.4 \pm 2.7$           & $\mathbf{73.1 \pm 0.4}$  & $47.0 \pm 3.1$   \\
            \cmidrule{1-8}
            \multirow{8}{*}{\rotatebox{90}{Self-supervised}}&BT + SRN  & $59.6 \pm 1.3$           & $56.7 \pm 1.4$           & $66.9 \pm 0.6$           & $66.1 \pm 0.5$            & $58.3 \pm 0.3$           & $57.6 \pm 0.3$   \\
            &BT + SRN + Size                                           & $63.3 \pm 0.8$           & $63.7 \pm 0.7$           & $68.3 \pm 0.8$           & $65.8 \pm 0.4$            & $\mathbf{59.9 \pm 0.3}$  & $\mathbf{59.1 \pm 0.2}$   \\
            &BT + SRN + SD-CL                                          & $61.8 \pm 2.6$           & $60.4 \pm 2.1$           & $67.3 \pm 0.9$           & $\mathbf{66.7 \pm 0.5}$   & $58.5 \pm 0.7$           & $59.0 \pm 0.5$   \\
            &BT + SRN + SD-CL  + Size                                  & $\mathbf{64.3 \pm 1.1}$  & $\mathbf{64.2 \pm 1.4}$  & $\mathbf{68.4 \pm 0.5}$  & $65.7 \pm 0.5$            & $\mathbf{59.9 \pm 0.3}$  & $59.0 \pm 0.2$   \\
            \cmidrule{2-8}
            &MoCo + SRN                                                & $49.9 \pm 7.9$       & $53.5 \pm 5.7$      & $36.0 \pm 1.8$           & $10.8 \pm 5.0$     & $35.4 \pm 8.0$  & $37.0 \pm 4.0$   \\
            &MoCo + SRN + Size                                         & $55.3 \pm 0.3$       & $39.9 \pm 10.7$     & $37.3 \pm 0.8$           & $26.5 \pm 3.9$     & $40.8 \pm 1.8$  & $30.6 \pm 4.0$   \\
            &MoCo + SRN + SD-CL                                        & $55.2 \pm 4.4$       & $38.6 \pm 8.0$      & $37.2 \pm 1.5$           & $26.7 \pm 3.7$     & $41.2 \pm 1.5$  & $32.1 \pm 3.9$   \\
            &MoCo + SRN + SD-CL + Size                                 & $54.7 \pm 2.0$       & $38.3 \pm 9.5$      & $37.5 \pm 0.9$           & $25.8 \pm 4.2$     & $39.5 \pm 2.3$  & $31.6 \pm 3.1$   \\
            \bottomrule
         \end{tabular}
         }
         \caption{Comparison of different methods for nuclei encoding on TNBC, CoNSep and PanNuke datasets. We bold the best score in its category, self-supervised learning and supervised. Our proposed convolution layer is named SD-CL. SRN is a modified 34-ResNet adapted for small images \cite{he2016deep}. MoCo \cite{he2020momentum} and BT \cite{zbontar2021barlow} have not yet been applied to nuclei encoding in histopathology images. $L$ is the number of classes, $n_{train}$ and $n_{test}$ are the number of samples in the train and test sets.}
         \label{tab:results}
      \end{center}
      \vspace{-.15in}
 \end{table*}

\vspace{.05in}\noindent {\bf Results}
We run our experiments on a High-performance computing cluster and repeat each method $20$ times. 
We display the average accuracy and standard deviations for the different methods in Table \ref{tab:results}.

The manual designed features achieve scores at least as well as the pre-trained ResNet except for the kNN accuracy on CoNSeP.
We notice that injecting the size to the pre-trained ResNet feature is detrimental for the encoding as the kNN accuracy loses more than $20\%$ for TNBC and PanNuke.
It is possible that the scale of the added feature penalises the encoding, the scale would not penalise the linear accuracy as the feature can be weight and the scales can be learned and adapted to the situation.
Generally speaking the models where no training occur under-perform compared to the learning paradigms.
We can however imagine a situation, when the number of data points is very low, that these features work reasonably well.

When the number of samples is low, i.e. for the TNBC dataset, the self-supervised framework outperforms the supervised one with a relative difference of $5.5\%$ for the linear accuracy and $5.3\%$ for the kNN accuracy.
In comparison, on the CoNSep and PanNuke dataset, the supervised methods outperform the self-supervised method by $6.0\%$ and $13.2\%$ on the linear accuracy and by $4.9\%$ and $5.8\%$ on the kNN accuracy.
This behaviour is expected as the supervised models need more data in order to reach a correct representation of the nuclei.

Injecting the size information into the latter layers prior to the encoding generally boosts the linear score and lowers the kNN score.
Similarly to the pre-trained ResNet this could simply be a question of scale and correct weighting.
Apart for the CoNSep data, the BT model is the only one where both linear and kNN scores increase with the size injection.   
In particular, the best performing self-supervised models are reached when the size is present, except for CoNSep on kNN score.

For self-supervised nuclei encoding, BT should be prioritised over MoCo.
Indeed the simpler formulation and the difference in linear and kNN score proves the effectiveness of BT over MoCo in these datasets.
In addition, to boost MoCo's performance, we increased the batch size to $1024$ as well as increased the learning rate range during the model selection phase.

Finally, the SD-CL achieves at least as well as the standard convolution layer.
For most datasets and most situations, i.e. supervised, self-supervised and with/without size, the effect of the SD-CL layer is positive.
The difference in performance is more acute when the sample size is small.
Theoretically, it should allow features to be extracted better from the images.
However, with enough samples, we suspect that the model can learn clues regarding the original scale of the image. 
As images are resized using a bi-linear interpolation, smaller images will show smooth color gradients whereas larger images will be sharper.

\vspace{.05in}\noindent {\bf Weight visualisation: } In Fig. \ref{fig:weights}, we visualise the weights of the first layer of SResNet and of SResNet + SDCL model. 
For visualisation purposes, we perform the min-max normalisation on the whole set of kernels and rescale the range to $[0,255]$ and plot the resulting RGB weights.
From the figure, both sets of weights are reasonable and conform with the literature \cite{zeiler2014visualizing}, such as the presence of a homogeneous colour or colour gradients.
Expectedly, we notice the presence of purple and pink filters matching the H\&E staining.

\insertweighs

\section{Conclusion} \label{conclusion}

We compare hand-designed features, pre-trained models, supervised models and self-supervised models on three nuclei type datasets that range from small to large samples size.
In this benchmark, we show the effectiveness of differently produced encodings.
The self-supervised learning Barlows-twins model ahead in the low sample setting and the supervised learning ahead in the large sample setting.
Moreover, we presented a new convolutional layer for extracting features dependant on the scale.
In addition to accounting for the scale, this layers impacts positively the performances and shows similar weight introspection to the standard convolution layer.
In the spirit of open science and reproducibility, the cell type extension dataset for TNBC and the code are made publicly available.

\section*{Data and code availability}
The TNBC segmentation dataset is available at \url{https://zenodo.org/record/1175282}.
The extended TNBC classification dataset is available at \url{https://zenodo.org/record/3552674}.
The CoNSeP segmentation and classification dataset is available at \url{https://warwick.ac.uk/fac/cross_fac/tia/data/hovernet/}.
The PanNuke segmentation and classification dataset is available at \url{https://warwick.ac.uk/fac/cross_fac/tia/data/pannuke}.
The code for MoCo used the lightly package \url{https://docs.lightly.ai/index.html} and the code for BT was inspired from \url{https://github.com/yaohungt/Barlow-Twins-HSIC}.
In addition methodological developments, for the sake of reproducible research and open science, we make the code for all experiments publicly available in the following Github repository \url{https://github.com/PeterJackNaylor/ScaleDependantCNN}.

\section*{Acknowledgement}
We thank Dr. Thomas Walter for his support in the project and in particular for the insightful discussions, help with the annotations and use of the software.
M.Y. was supported by MEXT KAKENHI 20H04243 and partly supported by MEXT KAKENHI 21H04874. We are also thankful for the RAIDEN computing system and its support team at the RIKEN AIP, which we used for our experiments. 

\newpage
{\small
\bibliographystyle{ieee}
\bibliography{main}
}

\end{document}